# Non-Monotonicity in Probabilistic Reasoning

Benjamin N. Grosof*

Computer Science Department, Stanford University
Building 460, Stanford, California 94305
(415) 723-4638 ; Arpanet: *grosof@score.stanford.edu*

## Abstract

We start by defining an approach to non-monotonic probabilistic reasoning in terms of non-monotonic categorical reasoning. We identify a type of non-monotonic probabilistic reasoning, akin to default inheritance, that seems to be commonly found in practice. We formulate this in terms of the *Maximization of Conditional Independence (MCI)*, and identify a variety of applications for this sort of default. We propose a formalization using *Pointwise* Circumscription. We compare MCI to Maximum Entropy, another kind of non-monotonic principle, and conclude by raising a number of open questions.

## 1 Introduction

Probabilistic reasoning is often rather loosely referred to as being non-monotonic. But how can we make this notion of non-monotonicity precise? In particular, what sort of non-monotonicity characterizes probabilistic reasoning? We will be addressing both of these questions in this paper. Our approach is to use the same definition of non-monotonicity as for categorical reasoning. This requires that we cast probabilistic reasoning in terms of categorical reasoning, i.e. reasoning where each sentence may only take on a truth value of 0 or 1.

## 2 Probabilistic Logic

A way to do so is provided by the approach of "Probabilistic Logic", as introduced in [15] and extended in [3].

Suppose we start with a (finite) set $\mathcal{A}$ of statements about the probabilities of some sentences $S_i$ in a (categorical) logical language $\mathcal{L}_b$. The basic idea of Probabilistic Logic is to express the probabilistic statements $\mathcal{A}$ in a second logical language $\mathcal{L}_m$ which is classical and categorical, e.g. first-order. The formulas of $\mathcal{L}_b$ are *reified*: they become terms in $\mathcal{L}_m$. The connectives of $\mathcal{L}_b$ become functions in $\mathcal{L}_m$. The logical properties of $\mathcal{L}_b$ (e.g. its tautologies) are encoded as a set of axioms $\mathbf{LOG_b}$ in $\mathcal{L}_m$. We can represent an assertion $G$ in $\mathcal{L}_b$ by an axiom $P(G) = 1$ in $\mathcal{L}_m$ which says it holds with certainty. For example we can write as an axiom in $\mathbf{LOG_b}$:

$$\forall XY.P((X \wedge (X \rightarrow Y)) \rightarrow Y)) = 1$$

*This work was supported by the author's National Science Foundation Graduate Fellowship; and by the Defense Advanced Research Projects Agency, the Office of Naval Research, and Rockwell International under contracts N00039-83-C-0136, N00014-81-K-0004, and B6G3045.

There is a subtlety, however, that eases the job of writing $\mathbf{LOG_b}$. The only formulas of $\mathcal{L}_b$ that we need to describe in $\mathcal{L}_m$ are those formed (by application of logical connectives) from the members of the set $\mathcal{F}$ of interpretation classes of $\mathcal{S} = \{S_i\}$. Then we do not need to encode in $\mathbf{LOG_b}$ all the tautologies of $\mathcal{L}_b$, only those of propositional logic involving propositions corresponding to the space $\mathcal{W} = 2^{\mathcal{F}}$. See [15] for details.

The definitional axioms of standard probability $\mathbf{AXPROB}$, are also encoded in $\mathcal{L}_m$, e.g. [1]:

$$P(True) = 1$$

$$\forall XY.\{P(\neg(X \wedge Y)) = 1\} \Rightarrow \{P(X \vee Y) = P(X) + P(Y)\}$$
$$\forall XY.\{P(Y) \neq 0\} \Rightarrow \{P(X|Y) = \{P(X \wedge Y)/P(Y)\}\}$$
$$\forall XY.\{P(X \equiv Y) = 1\} \Rightarrow \{P(X) = P(Y)\}$$

From these we can get the usual probabilistic identities, including Bayes' Theorem.

Thus by recourse to a meta-language, probabilistic knowledge and reasoning can be described in terms of categorical logic and reasoning.

**Disclaimer:** There are some kinds of probabilistic reasoning, notably with "higher-order" probabilities, which cannot be straightforwardly formulated in the above way. That is, we have assumed that $\mathcal{A}$ contains statements in which there is only one level of nesting

---

[1] cf. [5], modified so that the probability function is defined on propositions rather than sets

of {probability of ...} before reaching a proposition in a categorical language. However, one could apply our approach iteratively to express some sorts of second-order or higher-order probabilistic theories. Above, we assumed $\mathcal{L}_b$ was zero-th order in $P$. Then $\mathcal{L}_m$ was first-order in $P$. However, more generally, we could let $\mathcal{L}_b$ be of order $n$ in $P$, and then $\mathcal{L}_m$ would be of order $n+1$.

## 3 Non-Monotonic Probabilistic Theories

Let $\mathbf{Th}(\mathcal{Q})$ denote the closure of $\mathcal{Q}$ under logical implication in $\mathcal{L}_m$. Furthermore, let

$$\mathbf{Th^a}(\mathcal{A}) \triangleq \mathbf{Th}(\mathcal{A} \cup \mathbf{LOG_b} \cup \mathbf{AXPROB})$$

Then $\mathbf{Th^a}(\mathcal{A})$ is what [15,3] call "the" *probabilistic theory* which is *probabilistically entailed* by $\mathcal{A}$. There $\mathcal{L}_m$ was implicitly first-order logic.

We say $\mathcal{A}$ is of *Type-1-ci* form (terminology of [3]) iff it consists of axioms of the form $P(S_{i1}|S_{i2}) \geq q_i$, where $q_i$ is a number in the real unit interval $[0,1]$. (Note that $P(X|Y) \leq r$ can be re-written equivalently as $P(\neg X|Y) \geq (1-r)$.) In such a case, $\mathbf{Th^a}(\mathcal{A})$ is equivalent to a *lower probability distribution* $\mathbf{P}^-(\mathcal{W}|\mathcal{W})$ giving lower (and upper) bounds on all the conditional probabilities defined on $\mathcal{W}$.

**Definition 3.1:** A set operator $\mathbf{T}$ is *monotonic* iff

$$\forall \mathbf{B}_1 \mathbf{B}_2.\{\mathbf{B}_1 \subseteq \mathbf{B}_2\} \Longrightarrow \{\mathbf{T}(\mathbf{B}_1) \subseteq \mathbf{T}(\mathbf{B}_2)\}$$

**Definition 3.2:** A set operator $\mathbf{T}$ is *non-monotonic* iff it is not monotonic, i.e. iff

$$\exists \mathbf{B}_1 \mathbf{B}_2.\{\mathbf{B}_1 \subseteq \mathbf{B}_2\} \wedge \{\mathbf{T}(\mathbf{B}_1) \not\subseteq \mathbf{T}(\mathbf{B}_2)\}$$

Probabilistic Logic allows us to consider $\mathcal{A}$ to be a set of statements in $\mathcal{L}_m$. Suppose we have some rule for generating the set of conclusions (in $\mathcal{L}_m$) that we draw from $\mathcal{A}$. We can then ask whether the set of conclusions grows monotonically as we add to $\mathcal{A}$. We call this conclusion-drawing set operation *theory-closure*. $\mathbf{Th^a}$ above is an example of a theory-closure operator. *Logical (non-)monotonicity* is (non-)monotonicity of theory-closure. If $\mathbf{T}$ is (non-)monotonic, we say that $\mathbf{T}(\mathbf{B})$ is a (non-)monotonic theory.

**Definition 3.3:** A sentence $C$ in a non-monotonic theory $\mathbf{T}(\mathbf{B_1})$ is *defeasible* iff

$$\exists \mathbf{B}_2.\{\mathbf{B}_1 \subset \mathbf{B}_2\} \wedge \{C \notin \mathbf{T}(\mathbf{B}_2)\}$$

We also call a defeasible conclusion a *non-monotonic conclusion*. A monotonic conclusion is one which is not defeasible. Thus we can partition any theory $\mathbf{T}(\mathbf{B})$ into a monotonic part and a non-monotonic part. We will say that *reasoning* (inference) is monotonic or non-monotonic according to whether the conclusions drawn are monotonic or non-monotonic. We write $\models_\mathbf{T}$ for entailment of a monotonic conclusion, and $\approx_\mathbf{T}$ for entailment of a non-monotonic conlusion.

**Definition 3.4:** An *update* $\Delta$ to $\mathbf{T}(\mathbf{B})$ is *non-monotonic* iff $\mathbf{T}(\mathbf{B}) \not\subseteq \mathbf{T}(\mathbf{B} \cup \Delta)$.

Note that when $\mathbf{T}(\mathcal{A})$ is of Type-1-ci form, logical monotonicity corresponds to the functional monotonicity of the equivalent $\mathbf{P}^-(\mathcal{W}|\mathcal{W})$. In other words, if when we add new probabilistic statments to $\mathcal{A}$, the lower bound of every probability does not decrease, then the update is monotonic. If this condition is violated, then the update is non-monotonic.

Probabilistic Logic as defined in [15,3] draws its notions of logical implication and entailment from classical categorical logic, in fact implicitly from first-order logic. It is thus monotonic, since first-order logic is.

However, just as classical logic can be extended to perform non-monotonic reasoning, so can Probabilistic Logic. Thus we can represent non-monotonic probabilistic reasoning via non-monotonic categorical reasoning in $\mathcal{L}_m$. To do so will require us to adopt theory-closure operators different from $\mathbf{Th^a}$, since $\mathbf{Th^a}$ is monotonic. However, we might want to include $\mathbf{Th^a}$ in the monotonic part of whatever non-monotonic $\mathcal{T}$ we employ. Later when we define a circumscriptive approach to one kind of non-monotonic probabilistic reasoning, we will do just that. We will call the monotonic sentences (e.g. $\mathbf{Th^a}(\mathcal{A})$) *hard* information (beliefs). Relatively speaking, the non-monotonic conclusions are *soft*, i.e. tentative.

*Non-monotonic probabilistic reasoning requires us to employ principles for drawing conclusions which properly extend (i.e. go beyond) the axioms of classical probability.* Thus each of the types of non-monotonic probabilistic reasoning discussed below takes the axioms of a classical probability as a constraining point of departure rather than as an equivalent model.

### 3.1 A Monotonic Example

As an example of monotonic probabilistic reasoning using $\mathbf{Th^a}$ as our theory-closure operator, consider the case of a rather rowdy fellow named Igor. Let *Fights* denote the proposition that Igor gets in a bar fight; let *Drunk* denote the proposition that Igor has more than three drinks. Suppose to begin with we are given (i.e. $\mathcal{A}_1$ consists of):

$$P(Fights|Drunk) = .6$$
$$P(Drunk) = .3$$

Then we can infer

$$\mathcal{A}_1 \models_{\mathbf{Th}^a} \{P(Fights) \geq .18\} \quad (1)$$

If we next learn (i.e. add to $\mathcal{A}_1$ to get $\mathcal{A}_2$)

$$P(Fights|\neg Drunk) = .2$$

then we can infer

$$\mathcal{A}_2 \models_{\mathbf{Th}^a} \{P(Fights) = .32\} \quad (2)$$

which is consistent with, but stronger than, (1). The conclusions (1) and (2) are forced or determined by the given information in a strong sense which depends only on the standard axioms and definitions of classical probability.

### 3.2 A Non-Monotonic Example

Next we investigate an example of what appears to be one important type of non-monotonic probabilistic reasoning. Consider the case of 1985 model-year cars made by Neptune Corporation. Let $L$ denote the proposition that a car is so severely defective that it mechanically breaks down in its first 1000 miles. Let $N$ denote the proposition that a car's maker is Neptune. To begin with we are given (i.e. $\mathcal{A}_1$ consists of)

$$P(L|N) = .1$$

(No wonder you have never heard of Neptune Corporation before.) Suppose we are now asked what is $P(L|(N \wedge T))$, where $T$ means the car is a Triton model. $\mathbf{Th}^a(\mathcal{A}_1)$ tells us nothing about the value of $P(L|(N \wedge T))$: it could consistently be anything between 0 and 1.

However, a commonly-found pattern of probabilistic reasoning is to presume in this circumstance that

$$P(L|(N \wedge T)) = P(L|N)$$

A variety of rationales might be offered. One is that as long as we have no information to the contrary, the best estimate of the proportion of lemons in the class of Neptune Tritons is to use the information we are given about the proportion of lemons in the overall class of Neptunes. Another rationale is that since we have no evidence that the property of being a Triton model is indeed relevant to whether Neptunes are lemons, we will presume it is irrelevant.

In effect,
$$P(L|(N \wedge T)) = .1$$
is adopted as a *default*. We are using a non-monotonic theory-closure operator $T$ to generate a non-monotonic conclusion:

$$\mathcal{A}_1 \models_T \{P(L|(N \wedge T)) = .1\} \quad (3)$$

Suppose next we learn (i.e. add to $\mathcal{A}_1$ to form $\mathcal{A}_2$)

$$P(L|(N \wedge T)) = .05 \quad (4)$$

i.e. we get definitive, hard information about the value of $P(L|(N \wedge T))$. (4) contradicts and overrides our previous non-monotonic conclusion (3): it is a non-monotonic update. If next we are asked what is $P(L|(N \wedge T \wedge W))$, where $W$ means that the car is a station-wagon, our circumstance is similar to that above. Again, $\mathbf{Th}^a$ tells us nothing: only that $P(L|(N \wedge T \wedge W))$ may consistently take on any value between 0 and 1. Later, as we did in (4), we may get specific, hard information. In the meanwhile, we might apply the same sort of non-monotonic reasoning as we performed before to get (3). This time there is an added complexity, though. We have two different pieces of hard information bearing on the probability of $L$: both are conditioned on classes which are more general than $(N \wedge T \wedge W)$. Often a refinement to the above rationales is invoked: in cases of such competition, we choose to adopt the "most specific" information, i.e. the one which is conditional on the most specific class. So in the choice between

$$P(H|(N \wedge T \wedge W)) = P(H|N); \text{and} \quad (5)$$
$$P(H|(N \wedge T \wedge W)) = P(H|(N \wedge T)) \quad (6)$$

we favor the latter. Thus we infer

$$\mathcal{A}_2 \models_T \{P(L|(N \wedge T \wedge W)) = .05\} \quad (7)$$

Similarly, if we are asked about the probability of L for progressively more specific classes (e.g. by adding blue, air-conditioned, etc. as further conditions), we might employ the same pattern of non-monotonic reasoning to conclude from $\mathcal{A}_2$ that:

$$P(L|(N \wedge T \wedge W \wedge Blue)) = .05,$$
$$P(L|(N \wedge T \wedge W \wedge Blue \wedge AirCond)) = .05, \ldots$$

## 4 Default Inheritance of Probabilities

Our non-monotonic example above illustrates what appears to be one commonly-found type of non-monotonic probabilistic reasoning. Now we will

93

formulate the example more abstractly. Define

$$H \triangleq L$$
$$C_1 \triangleq N$$
$$C_2 \triangleq (N \wedge T)$$
$$S \triangleq (N \wedge T \wedge W)$$
$$q_1 \triangleq .1$$
$$q_2 \triangleq .05$$

$\mathcal{A}_2$ consisted exactly of:

$$P(H|C_1) = q_1$$
$$P(H|C_2) = q_2$$

while **LOG$_b$** contained:

$$P(C_2 \to C_1) = 1$$
$$P(S \to C_1) = 1$$
$$P(S \to C_2) = 1$$

In our example, we *inherited* a defeasible (default) value for the probability of $H$ for the conditioning class $S$ from the most specific conditioning class $C_i$ for which we had a hard value for the probability of $H$.

**The "Default Inheritance" Principle:** Let $P(H|S)$ denote the probability of some hypothesis $H$ of interest, given the situation $S$ at hand. Suppose our hard information $\mathcal{A}$ consists only of values for the probability of $H$, conditional on various propositions $C_i$ which form a chain. Then in order to conclude a defeasible value for $P(H|S)$, we look for the most specific $C_j$ such that $S \to C_j$, and make $P(H|S)$ equal to $P(H|C_j)$.

The structure of this sort of non-monotonic reasoning is analogous to that of default inheritance in categorical reasoning, e.g. in the classic example of whether birds and ostriches fly. In default inheritance, a particular class $S$ inherits an attribute $A$ from the most specific class $C_j$ of a chain of $S$'s superclasses $\{C_i\}$ for which information about $A$ is available. In the categorical case of default inheritance, the attribute is inherited with certainty, e.g. *Flies* or else *¬Flies*. We can represent this as inheriting either $P(A) = 0$ or else $P(A) = 1$[2]. The "default inheritance" type of non-monotonic reasoning with probabilities corresponds to inheriting the probability $P(A)$ of the attribute, which is not always 0 or 1.

---

[2] We have considered here only binary attributes, but the property of certainty holds for n-ary attributes as well

Thus it can be formalized as a generalization of the usual default inheritance. Alternatively, we can think of it as inheriting with certainty an attribute which is a probability, e.g. $P(H)$ above.

## 5 Specificity-Prioritized Maximization of Conditional Independence

We can formulate our non-monotonic example in terms of (non-monotonically inferred) conditional independence statements. Let $CIG(\{x, y\}, z)$[3] mean that the propositions $x$ and $y$ are conditionally independent given the proposition $z$, i.e. that:

$$P((x \wedge y)|z) = P(x|z)P(y|z)$$

which is equivalent (when $P(x|z) \neq 0$) to:

$$P(x|(y \wedge z)) = P(x|z)$$

; and (when $P(y|z) \neq 0$) to: $P(y|(x \wedge z)) = P(y|z)$
Since $P((S \to C_i)) = 1$ for $i = 1, 2$ :

$$P(S \equiv (S \wedge C_i)) = 1$$

Thus (5) is equivalent to:

$$P(H|(S \wedge C_1)) = P(H|C_1); \text{i.e., } CIG(\{H, S\}, C_1).$$

Similarly, (6) is equivalent to:

$$P(H|(S \wedge C_2)) = P(H|C_2); \text{i.e., } CIG(\{H, S\}, C_2)$$

Note that after (4), because $P(H|C_2)$ differs from $P(H|C_1)$, $CIG(\{H, S\}, C_2)$ and $CIG(\{H, S\}, C_1)$ cannot hold simultaneously. In effect, we have a competition and conflict between the two. According to the "default inheritance" principle, we try to non-monotonically conclude at least one of the two, and $CIG(\{H, S\}, C_2)$ takes precedence when (as after (4), though not before (4)) there is conflict. Thus we can formulate the precedence of more specific information as a *priority* among default conditional independence statements.

We *propose* formulating the "default inheritance" principle as the *Specificity-Prioritized Maximization of Conditional Independence (SPMCI)*. That is, given some hard probabilistic axioms, we non-monotonically conclude conditional independence statements corresponding to inheritance chains. (If such conditional independence statements are inconsistent with the given hard axioms, then as usual with

---

[3] Here $\{x, y\}$ is a set not a tuple, since $CIG(\{x, y\}, z)$ is symmetric in $x \leftrightarrow y$. It is also useful to define the case of mutual independence among a set of $n$ propositions conditional on $z$, but we will not take the space here.

94

defaults we block them as conclusions.) In case of the sort of conflict above, we apply precedence based on specificity in the above sense.

An important (and open) question is which conditional independence statements to maximize. We may only want to apply the "default inheritance" principle to some hypotheses $H^k$ and some chains $C_i^l$ and situations $S^m$. If we are only interested in inheriting a default value the way we did above for the probability of a particular $H$ conditional on a particular $S$, then it appears we need consider only $CIG$ tuples $\langle \{u,v\}, w \rangle$ such that $H$ or $\neg H$ is in $\{u,v\}$ and $P(S \to w) = 1$.

## 6 Non-Monotonicity in "Evidential" Reasoning

An important type of probabilistic reasoning in AI has been what we will call "evidential" reasoning, in which we are given hard information about $P(H|E_i)$ for each of several $E_i$'s. Importantly, no $E_i$ subsumes any other, though they may (and ususally do) overlap. They do not form a chain. By making the assumptions of conditional independence of the $E_i$'s given both $H$ and given $\neg H$, one can then infer a value for

$$P(H|(E_1 \wedge \ldots \wedge E_n))$$

(E.g., see PROSPECTOR [2], as well as MYCIN and Dempster-Shafer [4].)

### 6.1 "Default Inheritance"

Typically, this value for $P(H|(E_1 \wedge \ldots \wedge E_n))$ is in effect combined (rather implicitly) with the "default inheritance" principle to yield non-monotonically a value for $P(H|S)$, when ($S$ is the situation at hand and) we believe with certainty that:

$$S \to (E_1 \wedge \ldots \wedge E_n)$$

and when $S$ implies no other $E_j$'s for which $P(H|E_j)$ is available. This step corresponds to an application of SPMCI; more specific conjunctions of the $E_i$'s take precedence.

### 6.2 "Soft-Coding" Assumptions

A problem with "evidential" reasoning schemes is that the conditional independence assumptions of the $E_i$'s given $H$ and given $\neg H$ are often too strong: there are so many such assumptions that they are inconsistent either with each other, or with given (hard) information about the probability of $H$ given conjunctions of various $E_i$'s. We observe that MCI can be used to make such assumptions by default. Past approaches have been to "hard-code" or "build" such assumptions into the probabilistic inference machinery in a way which is monotonic and thus frequently inconsistent. "Soft-coding" via defeasibility retains the advantages (conceptual simplicity, representational parsimony, and computational ease) afforded by making the assumptions, to the greatest extent possible without sacrificing consistency and expressiveness. We can regard this as maximizing, rather than inflexibly assuming, a sort of "modularity" or "locality".

Another issue in evidential reasoning is that often $P(E_i|S)$ is uncertain rather than certain. In such cases, commonly (e.g. in PROSPECTOR [7]) the assumption is made that for each of several $EJ_k$ representing most specific conjunctive formulae in the $E_i$'s and their negations:

$$P(H|(EJ_k \wedge S)) = P(H|EJ_k)$$

i.e. that:

$$CIG(\{H, S\}, EJ_k)$$

Of course this assumption may be inconsistent with other hard information. In particular, the presumption that it is consistent in practice seems to have been made by implicitly limiting what sorts of probabilistic information will be present, i.e. can be expressed, in the AI system making this assumption [8]. If we "soft-code" this assumption as a "default inheritance" step, then we can avoid the choice between expressive limitation and inconsistency.

## 7 Graphoids, Influence Diagrams, and Irrelevance

Recently both the AI and the Decision Analysis research communities have developed interest in the idea of reasoning about the structure of (conditional) dependencies and independencies among a complexly-related collection of probabilistic events, in a fashion abstracted from the details of the particular underlying probabilistic values It appears that especially for humans it is a natural and helpful way to factor probabilistic reasoning. This makes it important for explanation, justification, and validation of probabilistic reasoning, and suggests that there may be computational advantages as well.

One direction of this research is represented by *influence diagrams* [6]. Influence diagrams implicitly specify conditional independencies by omission of "links" representing conditional probability statements (constraints). This suggests the use of a *non-monotonic specification convention* for influence diagrams: a sort of "closed dependency" assumption analogous to the "closed world assumption" familiar in categorical reasoning.

A related direction of research is the alternative formulation of conditional independence provided by the abstraction of Graphoids [16][4]. A Graphoid is the theory of a trinary relation, $I(x,z,y)$, which we can take to denote $CIG(\{x,y\},z)$, but with the additional generality that $x$, $y$, and $z$ denote (non-intersecting) *sets* of propositions. Informally,

$$I(\{a_1,\ldots,a_l\},\{c_1,\ldots,c_m\},\{b_1,\ldots,b_n\})$$

denotes

$$\bigwedge_{i,j,k} I(\{a_i\},\{c_k\},\{b_j\})$$

; and $I(\{a_i\},\{c_k\},\{b_j\})$ denotes $CIG(\{a_i,b_j\},c_k)$.

As with influence diagrams, we can imagine employing non-monotonic reasoning about Graphoids, e.g. as a specification convention. MCI in terms of Graphoids is the maximization of the $I$ relation. Thus given partial constraints on the relation $I$, we might non-monotonically conclude additional positive literals in $I$.

Another way to think about conditional (in)dependence is in terms of *(ir)relevance*. $CIG(\{x,y\},z)$ means that given $z$, learning $y$ is irrelevant to our estimate of the probability of $x$; and vice versa, that given $z$, learning $x$ is irrelevant to our estimate of the probability of $y$. MCI then corresponds to the non-monotonic *maximization of irrelevance*. This has a flavor of maximizing simplicity in the sense of Occam's Razor. The more that we can decide is irrelevant to some problem-soving task, the easier that task becomes; thus maximization of irrelevance holds out the ultimate promise of substantial computational advantages if that maximization itself is not too complex.

## 8 A Circumscriptive Formalization of (SP)MCI

We can try to formalize MCI in a variety of formalisms for non-monotonic reasoning, e.g. Circumscription [12,13], Default Logic [17], or Non-Monotonic Modal Logic [14]. However, we also want to express the precedence of more specific information in the sense discussed earlier. For this purpose, a recently-developed version of Circumscription, called Pointwise Circumscription [11], is most apt. In it, we can conveniently express priorities among the various defaults corresponding to particular conditional independence statements.

We lack space to go into the details here of circumscription and its pointwise version. The following treatment is necessarily rather schematic.

---
[4]Below we follow their notation

Circumscription accomplishes non-monotonic reasoning from a base theory **B** by applying the (monotonic) theory-closure of classical *second*-order logic to **B** augmented by an additional second-order *circumscription axiom* which is formed from **B** according to a *circumscription policy* specifying the non-monotonic behavior. The circumscription axiom expresses the minimality of a predicate.

$$T^{circ}_{policy}(\mathbf{B}) \triangleq Th_2(\mathbf{B} \cup Circ(\mathbf{B}; policy))$$

We now sketch a *proposed* method to construct an appropriate **B** and **policy** to accomplish (SP)MCI. We are currently investigating a number of unresolved outstanding technical issues involved in proving that the following indeed accomplishes its intended effect.

Let $\mathcal{B}_0$ (e.g. $\{\mathcal{A}\} \cup \mathbf{LOG_b} \cup \mathbf{AXPROB}\}$) be our "base" theory consisting of given, monotonic (hard), probabilistic axioms (both certain and uncertain), e.g. $P(C_2 \to C_1) = 1$; $P(H|C_1) = .15$; etc.. $\mathcal{B}_0$ is in a first-order language $\mathcal{L}_m$.

In terms of pointwise circumscription, we can express MCI via the circumscription, i.e. minimization, of an introduced abnormality predicate $AB$ characterized by the following axiom which we add to $\mathcal{B}_0$ to form $\mathcal{B}$. (By employing a slight variant of circumscription, which we dub "hyperscription", in which predicates are maximized rather than minimized, we can actually avoid the need to introduce an $AB$ and the following axiom. We just maximize $CIG$ directly. However the following formulation will be easier for most readers to follow.)

$$\neg AB(\{x,y\},z) \Rightarrow CIG(\{x,y\},z)$$

We can express MCI via a pointwise circumscription axiom[11]:

$$C_{AB}(\mathcal{B}; AB/V_{AB}, P/V_P, CIG/V_{CIG})$$

This says that $AB$ is minimized in the theory $\mathcal{B}$, with the predicates $AB$ and $CIG$ and the function $P$ being variable respectively when (the newly-introduced predicates) $V_{AB}$, $V_{CIG}$, and $V_P$ hold.

Two interesting sorts of questions about MCI are: which tuples $\langle\{x,y\},z\rangle$ to try to presume by default; and with what priorities. Pointwise circumscription gives us a way to specify these in some detail. We can express via $V_{AB}$ both the delimitation of the scope of MCI, and the priorities among various conditional independence (CI) defaults.

$V_{AB}(\langle\{u,v\},w\rangle,\langle\{r,s\},t\rangle)$ means that when minimizing $AB$ (i.e. maximizing $CIG$) at tuple $\langle\{u,v\},w\rangle$, the tuple $\langle\{r,s\},t\rangle$ is variable. To specify that we want MCI to apply to a particular tuple $\langle\{a,b\},c\rangle$, we include in $\mathcal{B}$ the axiom:

$$V_{AB}(\langle\{a,b\},c\rangle,\langle\{a,b\},c\rangle) \qquad (8)$$

If (8) is absent from $\text{Th}_2(B)$, e.g. if its negation is present, then MCI will not apply to that tuple.

To specify that the CI default on tuple $\langle\{a,b\},c\rangle$ has greater priority than the CI default on tuple $\langle\{d,e\},f\rangle$, we include in $\mathcal{B}$ the axiom:

$$V_{AB}(\langle\{a,b\},c\rangle,\langle\{d,e\},f\rangle)$$

Thus we can write a general *Specificity-Prioritization Axiom*:

$$\forall C_1, C_2.(P(C_2 \rightarrow C_1) = 1) \Rightarrow$$
$$(\forall x, y. V_{AB}(\langle\{x,y\},C_2\rangle,\langle\{x,y\},C_1\rangle))$$

We can imagine specifying other kinds of prioritizations among CI defaults as well. We may wish to infer some CI defaults before considering others. We can do so by making the former have higher priority, i.e. be relatively "harder".

Thus in pointwise circumscription we can[5] express Maximization of Conditional Independence with and without Specificity-Prioritization ((SP)MCI), restricted to arbitrary collections of tuples and with arbitrary priorities among the CI defaults.

## 9 Maximum Entropy

So far we have discussed two major types of non-monotonic probabilistic reasoning: "default inheritance" (formulable as SPMCI) and default locality/irrelevance/$I_{Graphoid}$ (formulable as MCI). A third type of non-monotonic probabilistic reasoning is the use of the Maximum Entropy assumption, which has attracted considerable attention in the AI community (e.g. [9,7,1,15,3])

Maximum Entropy (ME) is a method of selecting a non-monotonic extension of given ("base") hard axioms $\mathcal{B}$. The base axioms are treated as a set of constraints on the maximization of the entropy of the joint probability distribution $\mathbf{P}(\mathcal{F})$:

$$-\sum_{F_i \in \mathcal{F}} P(F_i) log(P(F_i))$$

ME always produces a unique, *complete* extension. By "complete", we mean that every $P(W_i|W_j)$ has a unique single real value in the ME extension: the lower conditional probability distribution on $\mathcal{W}$ is equal to the upper probability distribution.

---
[5]see the caveat above

Intuitively, ME tries to "flatten" the joint distribution $\mathbf{P}(\mathcal{F})$. In the extremal case, i.e. if the base theory is empty, then the result of ME is that each $P(F_i)$ is the same as every other. This is sometimes called the *uniform prior*, or *LaPlacian prior*, assumption.

ME often non-monotonically entails a large number of conditional independence statements. It has some elegant properties in this regard. A well-known result [10] is the Product Extension Theorem, which partially characterizes the sorts of conditional independence statements produced by ME, in terms of propositional subspaces.

A natural question is the relationship between ME and (SP)MCI.

Clearly they are not in general identical. Consider the case of an empty base theory. Here ME entails a uniform distribution, while (SP)MCI entails only conditional independence constraints which are satisfiable by non-uniform distributions. Also, in general (SP)MCI does not entail a unique, complete extension: e.g. it may result in bounds on, rather than point values for, some probabilities.

An interesting open question we are investigating is how fully to characterize the sort of conditional independence statements produced by ME, including in relation to specificity-prioritization.

## 10 Discussion

(SP)MCI appears to represent several important kinds of non-monotonicity in probabilistic reasoning. SPMCI can express the commonly-found "default inheritance" principle. We can use MCI as a specification convention for Graphoids or influence diagrams. We can use MCI to maximize irrelevance in a particular sense. MCI also promises to provide a tool to specify the presumption of "locality" of updating in the sense of "evidential" reasoning. MCI overlaps substantially with Maximization of Entropy (ME). Compared to ME, it is a more precisely controllable assumption. It separates the assumption of maximizing conditional independence from the uniform prior assumption; ME conflates the two. SPMCI can yield a non-monotonic theory with bounds, not just point values, for probabilities. Moreover, we can specify in much greater detail the tuples to which to apply MCI and SP. Hopefully that this will carry over to more control and goal-directedness in computation as well. Current ME algorithms are global, numerical relaxation procedures which calculate the entire joint probability distribution. It is thus often impracticably costly to apply the ME assumption. An open challenge is to make any of these three types of non-monotonic probabilistic reasoning reasonably efficient. One important criterion we might want to impose is that the non-monotonic conclusions about *CIG* be definite.

As usual with non-monotonic reasoning, there are at least two sorts of intepretations or justifications for adopting a non-monotonic theory-closure principle. One is as a representational or specification convention. Another is as a conjectural decision rule. Due to lack of space, we have concentrated here more on the form rather than on the pragmatic substance of non-monotonicity in probabilistic reasoning. One interesting lead we are investigating is the basis in Bayesian statistical estimation and decision theory for what we have called the "default inheritance" principle.

## 11 Conclusion

Probabilistic Logic casts monotonic probabilistic reasoning in terms of monotonic categorical reasoning with probabilistic statements. We extended this approach, and cast non-monotonic probabilistic reasoning in terms of non-monotonic categorical reasoning. We identified a type of non-monotonic probabilistic reasoning, akin to default inheritance in categorical reasoning, that seems to be commonly found in practice. We formulated this as a principle: *Specificity-Prioritized Maximization of Conditional Independence (SPMCI)*. We then identified another interesting type of non-monotonic probabilistic reasoning, akin to default irrelevancy, and showed that it can be formulated and formalized in similar terms: as Maximization of Conditional Independence (MCI). We formalized (SP)MCI using *Pointwise* Circumscription, a recently developed variant of the circumscription formalism for (categorical) non-monotonic reasoning. We noted the Maximum Entropy assumption as a third type of non-monotonic probabilistic reasoning, and compared it to (SP)MCI.

## 12 Directions for Future Research

The main intent of this paper is to help to define and provoke an area of investigation. We have offered more conjectures and suggestions than answers.

Several open questions about (SP)MCI were mentioned in passing. When, i.e. for which tuples, do we want to do MCI? Are there additional sorts of prioritizations besides SP which are desirable or useful? (Our preliminary investigations indicate that it is often undesirable to perform indiscriminate MCI, and that prioritization beyond specificity is sometimes desirable.) When and to what extent does ME produce MCI? Insofar as ME produces MCI, is it compatible with SP? Are there ways to employ (SP)MCI in relatively efficiently in goal-directed computations, i.e. without computing the entire lower probability distribution non-monotonically entailed by SPMCI? Does our proposed circumscriptive formalization of (SP)MCI have its intended models?

## Acknowledgements

Thanks to Peter Cheeseman, Michael Genesereth, David Heckerman, Eric Horvitz, Vladimir Lifschitz, Nils Nilsson, Judea Pearl, Devika Subramanian, and the participants of the Stanford Logic Group MUGS seminar, for valuable discussions and encouragement.